\documentclass[letterpaper]{article} 
\usepackage{aaai2026}  
\usepackage{times}  
\usepackage{helvet}  
\usepackage{courier}  
\usepackage[hyphens]{url}  
\usepackage{graphicx} 
\usepackage[numbers]{natbib}  
\usepackage{caption} 
\usepackage{algorithm}
\usepackage{amsmath,amssymb,amsfonts}
\usepackage{algorithmic}
\usepackage{textcomp}
\usepackage{multirow}
\usepackage{array}

\usepackage{newfloat}
\usepackage{listings}
\DeclareCaptionStyle{ruled}{labelfont=normalfont,labelsep=colon,strut=off} 
\floatstyle{ruled}
\newfloat{listing}{tb}{lst}{}
\floatname{listing}{Listing}
\title{MVAN: Multi-View Attention Networks for Fake News Detection on Social Media}
\author{
   Shiwen Ni, Jiawen Li, and Hung-Yu Kao*
}
\affiliations{
    \textsuperscript{\rm 1}Department of Computer Science and Information Engineering, National Cheng Kung University,
    Tainan, 701401, Taiwan

}

\usepackage{bibentry}

\begin{document}

\maketitle

\begin{abstract}
Fake news on social media is a widespread and serious problem in today's society. Existing fake news detection methods focus on finding clues from Long text content, such as original news articles and user comments. This paper solves the problem of fake news detection in more realistic scenarios. Only source shot-text tweet and its retweet users are provided without user comments. We develop a novel neural network based model, \textbf{M}ulti-\textbf{V}iew \textbf{A}ttention \textbf{N}etworks (MVAN) to detect fake news and provide explanations on social media. The MVAN model includes text semantic attention and propagation structure attention, which ensures that our model can capture information and clues both of source tweet content and propagation structure. In addition, the two attention mechanisms in the model can find key clue words in fake news texts and suspicious users in the propagation structure. We conduct experiments on two real-world datasets, and the results demonstrate that MVAN can significantly outperform state-of-the-art methods by 2.5\% in accuracy on average, and produce a reasonable explanation.
\end{abstract}


\section{Introduction}
With the rapid development of social media platforms, such as Twitter, fake news can spread rapidly on the internet and affect people’s lives and judgment. On April 27, 2020, the president of the U.S.A Donald Trump said, “Fake news, the enemy of the people!” Those words indicate that fake news has been a serious social problem. Fake news refers to false statements and rumors on social media, including completely false information or gross misrepresentation of a real event. However, due to the limitations of expertise, time or space, it is difficult for ordinary people to separate fake news from the vast amount of information available online. Therefore, it is necessary to develop automated and auxiliary methods to detect fake news at an early stage. With the development of artificial intelligence (AI), many researchers have attempted to apply AI technology to automatically detect fake news.

Early research on automatic detection of fake news mainly focused on designing effective features from various information sources, including text content \cite{b1,b2,b3}, publisher’s personal information \cite{b1,b4} and communication mode  \cite{b5}--\cite{b7}. However, these feature-based methods are very time-consuming and labour-intensive. In addition, the performance of the model is very dependent on the quality of the artificial features, thus, the performance of this method is not ideal in most cases.

Driven by the success of deep neural networks, several recent studies \cite{b8,b9} have applied various neural network models to fake news detection. For example, a recurrent neural network (RNN) \cite{b10} is used to learn the representation of the tweet text on the posting timeline. Liu \cite{b8} modelled the propagation path as a multivariate time series and applied a combination of RNN and convolution neural network (CNN) to capture the changes of user characteristics along the propagation path. The main limitation of these methods is that they can only process sequence data but cannot process structured data, leading to the inability to properly simulate the real propagation structure.

We know that the dissemination structure of the news on social media can constitute a social network graph. Generally, tweets can be reposted by any other user. The structure of tweet propagation composed of retweet users is shown in Fig. 1. With the help of social media, a piece of Twitter news can be spread all over the world in a very short time. To capture the information hidden in the sequence text and the structured propagation graph at the same time, RNN and graph neural network (GNN) were used to process these two kinds of data. Meanwhile, to make the model have better learning ability and certain interpretability, two different attention mechanisms, text semantic attention and propagation structure attention, were added to RNN and GNN.

\begin{figure*}[t]
	\centering
	\includegraphics[width=0.88\linewidth]{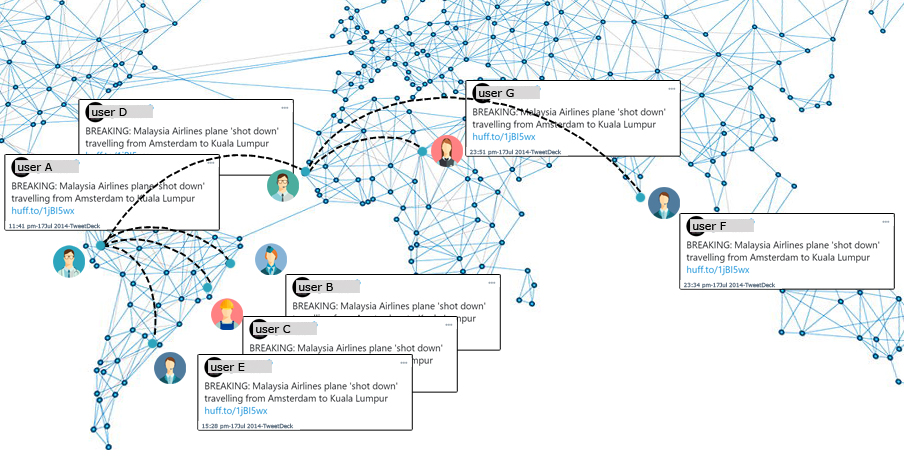}
	\caption{Propagation structure of Twitter on social media. A tweet can be reposted by multiple people, and a person can also repost multiple tweets.
	}
	\label{fig:f1}
\end{figure*}

The main contributions of this paper are summarized as follows:

(1) To the best of our knowledge, we are the first to adopt graph attention networks (GATs) to encode and represent the propagation structure of news.

(1) Experimental results on two real-world datasets show that the multi-view attention networks (MVAN) model achieves the highest accuracy and outperforms state-of-the-art models.

(3) Our model is more robust in early fake news detection and the model has some interpretability in both perspectives of text and propagation structure. 

\section{Related Work}
The goal of fake news detection is to distinguish the authenticity of news published on social media platforms based on their relevant information (such as text content, comments, propagation structure, etc.). Related works can be divided into different categories as follows.
\subsection{Feature-based methods}
Some early studies focused on fake news detection based on handcrafted features. These features are mainly extracted from text content and users’ profile information. Castillo et al. (2011) \cite{b1} proposed a decision tree-based model, utilizing a large number of features for fake news detection on Twitter. Yang et al. (2012) \cite{b4} created two new features to enrich the feature set of previous researchers: client-based features and location-based features. They are used to automatically detect fake news on Sina Weibo. Wu et al. (2015) \cite{b11} used a propagation structure composed of 23 features in the hybrid support vector machines (SVM). These features are divided into three categories (message-based features, user-based features and retransmission-based features). Kwon et al. (2017) \cite{b14} proposed a machine learning model based on time series fitting of tweet volume time characteristics. Ma et al. (2015) \cite{b10} proposed an SVM model that engineers each of the social context features. Rath et al. (2017) \cite{b15} extracted user information and combined them with an RNN model.

\subsection{Content-based methods}
Content-based methods rely on the text content to detect the truthfulness of the news article. Ma et al. (2015) \cite{b10} and Yu et al. (2016) \cite{b17} combined text and RNN or CNN for fake news detection. Chen et al. (2018) \cite{b18} combined the attention mechanism with the text to detect fake news at an early age. Liu (2018) \cite{b19} used both RNN and CNN to encode the propagation structure for early fake news detection. Ajao et al. (2018) \cite{b20}proposed a hybrid CNN-long-short term memory (LSTM) model for fake news detection on Twitter. Yu et al. (2019)  \cite{b21} proposed an attention-based convolutional approach for text authenticity detection. Shu et al. (2019) \cite{b22} proposed a sentence-comment co-attention sub-network to use both news contents and user comments for fake news detection and used an attention mechanism to provide explainability. Sujana et al. (2020)  \cite{b23}proposed a multi-loss hierarchical BiLSTM model for fake news detection, which encoded Twitter news from the post level and the event level. 
\subsection{Structure-based methods}
Unstructured methods cannot handle structured data. In recent years, researchers have proposed some new approaches to use structural information. Ma et al. (2018) \cite{b24} constructed a recursive neural network to handle conversational structure. This model generates a tree structure by bottom-up or top-down propagation. Monti et al. (2019) \cite{b25} proposed propagation-based fake news detection using graph convolutional networks (GCN). Nguyen (2019) \cite{b26} detected fake news using a multimodal social graph. Bian et al. (2020) \cite{b27} proposed a novel bi-directional graph neural networks model, bi-directional graph convolutional networks (Bi-GCN), to explore both characteristics by operating on both top-down and bottom-up propagation of fake tweets. Lu and Li (2020) \cite{b28} developed graph-aware co-attention networks (GCAN) to detect fake news, which generated an explanation by highlighting the evidence on suspicious re-tweeters and the words of concern. Li et al. (2020)  \cite{b29} built a conversation structure from source tweet and user comments, and used GNN to encode it. Li et al. (2020) \cite{b30}  crawled user-follower information and built a friendly network based on the follow-followers relationship.

We compare our work and the most relevant studies in Table 1. The uniqueness of our work lies in:
targeting at source news text, requiring no user response
comments, analysing model explainability and Integrating multiple attention mechanisms.
\begin{table}[]
	\renewcommand\arraystretch{1.2}   
	\caption{Comparison of recent related studies. Column notations:  source news texts (ST), response comments (RC), user features (UF), structural information (SI), model explainability (ME), attention mechanism(AT) and multiple attention mechanisms(MA).}
	\setlength{\tabcolsep}{2mm}{
		\begin{tabular}{|l|l|l|l|l|l|l|l|}
			\hline
			Works & ST & RC & UF & SI & ME & AT & MT \\ \hline
			BiLSTM \cite{b43} & \checkmark & \checkmark &  &  &  &  &  \\ \hline
			RCNN \cite{b46}  &\checkmark &  & \checkmark & \checkmark &  &  &  \\ \hline
			CSI \cite{b45}& \checkmark & \checkmark &\checkmark &  &  &  &  \\ \hline
			dEFEND \cite{b22} & \checkmark & \checkmark &  &  & \checkmark & \checkmark &  \\ \hline
			GCAN\cite{b28} & \checkmark &  &\checkmark & \checkmark &\checkmark & \checkmark &  \\ \hline
			G-SEGA\cite{b26} & \checkmark & \checkmark &  &\checkmark &  &  &  \\ \hline
			Our work & \checkmark&  & \checkmark & \checkmark& \checkmark &\checkmark & \checkmark \\ \hline
	\end{tabular}}
	\label{tab:my-table}
\end{table}
\section{The background of the related deep learning techniques}
Deep learning is a significant branch of machine learning. It has achieved unprecedented success in multiple natural language tasks, such as machine translation, emotion analysis, question answering system, etc. Deep learning is essentially a way of expressing learning that is different from traditional methods, which extracts features based on the manual method. Deep learning models can automatically generate appropriate vectors to represent words, phrases and sentences. This chapter focuses on relevant deep learning models used in this article. In this section, we introduce several deep learning techniques used in our model.
\subsection{Recurrent neural networks}
RNN is one of the most commonly used deep learning networks in natural language processing (NLP) tasks of AI. RNN is a type of feed-forward neural network that can be used to model variable-length sequential information such as sentences or time series. Therefore, it has some advantages in learning the nonlinear characteristics of sequences. For each time step, RNN updates its hidden state ht by extracting information from its last time step in a hidden state $h_{t-1}$ and the input in this time step $x_t$. This process continues until all time steps have been evaluated. The algorithm iterates over the following equations: 
\begin{equation}\label{key}
	a_t=f(h_{t-1},x_t)
\end{equation}
\begin{equation}\label{key}
	g(x)=tanhx
\end{equation}
\begin{equation}\label{key}
	a_t=g(\textbf{W}_{hh}\cdot h_{t-1}+\textbf{W}_{xh}\cdot x_t)
\end{equation}
\begin{equation}\label{key}
	a_t=tanh\textbf{W}_{hh}\cdot h_{t-1}+\textbf{W}_{zh}\cdot x_t
\end{equation}
\begin{equation}\label{key}
	h_t=\textbf{W}_{hy}\cdot a_t
\end{equation}
where $f$ and g are activation functions, $\textbf{W}_{hh}$ and $\textbf{W}_{xh}$ are parameters and $h$ means the hidden state, an RNN.
\subsection{Gated recurrent unit}
In practice, because of the vanishing or exploding gradients, the basic RNN \cite{b31} cannot learn long-distance temporal dependencies with gradient-based optimization (Bengio et al., 1994) \cite{b32}. One way to deal with this is to make an extension that includes “memory” units to store information over long periods, commonly known as LSTM unit (Hochreiter and Schmidhuber, 1997; Graves, 2013) \cite{b33,b34} and gated recurrent unit (GRU) (Cho et al., 2014) \cite{b35}. Here, we briefly introduce the GRU. Unlike an RNN unit, GRU has gating units that modulate the flow of the content inside the unit. The following equations are used for a GRU layer: 
\begin{equation}\label{key}
	z_t=\sigma (x_t\textbf{U}_z+h_{t-1}\textbf{W}_z)
\end{equation}
\begin{equation}\label{key}
	r_t=\sigma (x_t\textbf{U}_r+h_{t-1}\textbf{W}_r)
\end{equation}
\begin{equation}\label{key}
	\widetilde{h}_t=tanh(x_t\textbf{U}_h+(h_{t-1}\cdot r_t)\textbf{W}_h)
\end{equation}
\begin{equation}\label{key}
	h_t=(1-z_t)\cdot h_{t-1}+z_t\cdot \widetilde{h}_t
\end{equation}
where $\textbf{W}_z, \textbf{W}_r, \textbf{W}_h, \textbf{U}_z, \textbf{U}_r, \textbf{U}_h$ are parameters. $\sigma$ is a logistic sigmoid function.
\subsection{Word2vec}
Natural language is a special composition of characters produced by human beings for communication. For a computer to understand the natural language, words in the natural language need to be encoded. In the early days of NLP, words were often converted into discrete individual symbols according to the order in which the word appeared in a corpus. This encoding method was called one-hot-encoding. However, such an encoding method does not reflect the relationship between words. To overcome this problem, Bengio et al. (2003) \cite{b36} proposed the concept of word embedding for the first time in 2003. They assumed that each word in the glossary corresponds to a continuous eigenvector. This idea has since been widely applied to various NLP models, including word2vec. In word2vec, the two most important models are the CBOW model (continuous bag-words-model) and skip-gram model. The basic idea of CBOW and skip-gram is to make the vector represent the information contained in the word as fully as possible, while keeping the dimensions of the vector within a manageable range (between 25 and 1,000 dimensions, if appropriate). The CBOW model learns the expression of word vectors from the prediction of context to the target word. Mathematically, the CBOW model is equivalent to the embedding matrix of a word bag model multiplied by a successive embedding matrix. Conversely, the skip-gram model learns word vectors from the target word’s prediction of context. 
\subsection{Attention mechanism}

The attention mechanism, as the name suggests, is a technology that enables models to focus on important information. It is not a complete model but a technology that can be used in deep learning models. The mechanism was first proposed in the field of visual images. Mnih et al. (2014) \cite{b37} used the attention mechanism on the RNN model to classify images. Then, Bahdanau et al. (2014) \cite{b38} applied an attention mechanism to NLP that combined translation and alignment in machine translation tasks. In NLP, different words in a sentence have different importance, such as I hate this movie. In an emotional analysis, it is obvious that the word “hate” plays a more important role than the other words, which means that the model should concentrate on that word. Further, the attention mechanism is widely used in various NLP tasks based on neural network models, such as RNN/CNN. In 2017, the Google machine translation team made extensive use of the self-attention mechanism to learn textual representation \cite{b39}. The mechanism of self-attention has also become a recent research hotspot and has been explored in various NLP tasks.
\subsection{Graph attention networks}
Graph attention network (GAT) \cite{b41} is a kind of GNN. GNNs are especially deep learning-based methods that operate on non-Euclidean structures data domain. Due to their convincing performance and high interpretability, GNN is the widely applied graph analysis method recently \cite{b40}.

There are several variants of GNNs, of which GAT is the most commonly used. GATs are proposed by Velickovic et al. (2018) \cite{b41}. This kind of neural network incorporates the attention mechanism into the graph propagation step. It computes the hidden states of each node by attending over its neighbours, following a self-attention strategy. Velickovic et al. (2018) \cite{b41} proposed a single graph attentional layer and constructed arbitrary graph attention networks by stacking this layer. The layer computes the coefficients in the attention mechanism of the node pair (i, j) by: 
\begin{equation}\label{key}
	a_{ij}=\dfrac{\mathrm{exp(LeakyReLU}(a^T[\textbf{W} \vec{h}_i||\textbf{W}\vec{h}_j]))}{\sum_{k\in N_i}\mathrm{exp(LeakyReLU}(a^T[\textbf{W} \vec{h}_i||\textbf{W}\vec{h}_k]))}
\end{equation}
where $a_{ij}$ is the attention coefficient of node $j$ to $i$, $N_i$ represents the neighbourhoods of the $ith$ node in the graph. $\textbf{W}$ is the weight matrix of a shared linear transformation, which is applied to every node. $a$ is the weight vector of a single-layer feedforward neural network.
\section{Problem Statement}
Let $\Phi=\{s_1,s_2,...,s_{|\Phi|}\}$ be the set of source tweets (short-text) and $\Psi =\{g_1,g_2,...,g_{|\Psi|}\}$ be the set of propagation structure graph. Each source tweet $s_i\in \Phi$ corresponds to a propagation structure graph $g_i\in \Psi$. When a Twitter post $s_i$ is published, other users will retweet it. The propagation structure graph $g_i\in \Psi$ of each source tweet is composed of its retweet users $u_j$. $g_i=\{u_1,u_2,...,u_j\}$ and we denote $F=\{f_1,f_2,...,f_{|F|}\}$ as the set of user features. Given a source tweet $s_i$, along with the corresponding propagation $g_i$ containing users $u_j$ who retweet $s_i$, as well as their feature vectors $F_j$, the goal of our model is to classify $s_i$ as ‘true’ or ‘fake’. The classifier performs learning through labelled information, i.e., $C_i:s_i\rightarrow y_i$. In addition, we require our model to highlight some users $u_j\in  u_i$ who retweet $s_i$ and some words $q_i\in  s_i$ that can interpret why $s_i$ is predicted as a true or fake one.

\section{The proposed MVAN model}
We propose a novel neural network model, multi-view attention networks (MVAN), to detect fake news based on the source tweet and its propagation structure. MVAN consists of three components. The first is text semantic attention networks. Its role is to obtain the semantic representation of the source tweet text information. The second is propagation structure attention networks. It captures the hidden information in the propagation structure of a tweet. The last is the prediction module. It generates the final detection result by concatenating text semantic representation and propagation structure representation.
\subsection{Text semantic attention networks}
In this work, to correctly represent the information contained in the source tweet text and capture the key clue words in the source tweet text, we propose text semantic attention networks to process the source tweet text. We define $x_i\in \mathbb{R}^d$ as the d-dimensional word embedding corresponding to the $i$-th word in the source tweet. Because the length of each source tweet is different, we perform zero-padding here by setting a maximum length $L$. Let $E=[e_1,e_2,...,e_l]\in \mathbb{R}^l$ be the input vector of the source tweet, in which $e_l$ is the embedding vector of the word, when the ith word is the <pad> token, its embedding vector $e_l$ is an vector filled with 0. We use word2vec to encode the words of the source tweet. Moreover, a deep bi-directional gated recurrent unit (BiGRU) is used to capture the relationships among words and generate the source tweet representation:
\begin{equation}\label{key}
	h_t=\mathrm{BiGRU}(e_t,h_{t-1})
\end{equation}
where $h_t$ is the final hidden state of BiGRU, $h_t \in \mathbb{R}^{d_l}$. $e_t$ is the word embedding vector of the source tweet.

Since each word has a different role in detecting fake news, we believe that the model could focus on the keywords and reduce the role of other irrelevant words. We use a layer of the fully connected network to map the output vector $h_t$ of BiGRU to a matrix-vector $u_t$.
\begin{equation}\label{key}
	u_t=\mathrm{tanh}(\textbf{W}_wh_t+b_w)
\end{equation}
where $\textbf{W}_w \in \mathbb{R}^{n\times l}$ and $b_w$ are the weight and bias of attention and $\mathrm{tanh}$ is the activation function mapping the value between $[-1,1]$. $u_t \in \mathbb{R}^{n_l}$, and n is the number of neural units in the fully connected layer. Then we calculate the attention coefficient of each word, which is the final weight of each word.
\begin{equation}\label{key}
	a_t=\mathrm{softmax}(u_tu_w)=\dfrac{exp(u^T_tu_w)}{\sum _t exp(u^T_tu_w)}
\end{equation}
where $u_w$ is the weight matrix, $\cdot ^T$ represents transposition and $a_t \in  \mathbb{R}^l$.

Finally, each word vector $h_t$ and attention coefficient $a_t$ are weighted and summed to obtain the representation of the source tweet:
\begin{equation}\label{key}
	S=\sum_t a_th_t, S\in  \mathbb{R}^{o\times l}
\end{equation}
where $o$ is dimensional of output layer. Through text semantic attention networks, we get a representation vector containing text semantics and the attention weight of each word.
\subsection{Propagation structure attention networks}
Research shows that fake news and real news have different propagation structures \cite{b7}. Therefore, we propose propagation structure attention networks to capture the clues implicit in the propagation structure of news. In this part, we discuss how to encode the propagation structure into the node representation for fake news detection. Inspired by GAT \cite{b41}, we apply the attention mechanism to learn a distributed representation of each user node (retweet user) in the propagation structure graph by attending over its neighbours.

The input to the propagation structure attention layer is a set of user features, $u=\{\vec{u}_1,\vec{u}_2,...,\vec{u}_N\},\vec{u}_i\in  \mathbb{R}^F$, where $N$ is the number of user nodes and $F$ is the number of features in each user node. The output of the propagation structure attention layer is a new set of user node features, $p=\{\vec{p}_1,\vec{p}_2,...,\vec{p}_N\},\vec{p}_i\in  \mathbb{R}^{F'}$. Note that
$F'$ may not be equal to $F$.

To obtain enough presentation power to transform the original user features into higher-level features, at least one learnable linear transformation is required. For this purpose, as an initial step, a shared linear transformation, parametrized by a weight matrix, $\textbf{W}\in \mathbb{R}^{F'\times F}$, is applied to every user node. Then self-attention mechanism $att$ is used on each user node: $\mathbb{R}^{F'}\times \mathbb{R}^{F}\rightarrow \mathbb{R}$ computes attention coefficients that indicate the importance of user node $j$’s features to user node $i$.
\begin{equation}\label{key}
	c_{ij}=\mathrm{att}(\textbf{W}\vec{u}_i,\textbf{W}\vec{u}_j)
\end{equation}
We inject the propagation structure into the mechanism by performing masked attention, we only compute $c_{ij}$ for nodes $j\in U_i$, where $U_i$ is some neighbourhood of user node i in the propagation structure graph. In our experiment, each user node calculates the attention coefficients of all its first-order neighbours. To prevent the information of user node i from being forgotten, we regard user node i as its first-order neighbour. To make the attention coefficients easy to compare between different nodes, we use the softmax function to normalize them among all the choices of j:
\begin{equation}\label{key}
	a_{ij}=\mathrm{softmax}(c_{ij})=\dfrac{\mathrm{exp}(c_{ij})}{\sum _{k\in U_i}\mathrm{exp}(c_{ij})}
\end{equation}
In this paper, the propagation structure attention mechanism is a fully connected layer, parametrized by a weight vector $\vec{a}\in  \mathbb{R}^{2F'}$, using the LeakyReLU function nonlinearity. Fully expanded out, the coefficient calculated by the attention mechanism can be expressed as:
\begin{equation}\label{key}
	a_{ij}=\dfrac{\mathrm{exp(LeakyReLU}(\vec{a}^T[\textbf{W} \vec{u}_i||\textbf{W}\vec{u}_j]))}{\sum_{k\in U_i}\mathrm{exp(LeakyReLU}(\vec{a}^T[\textbf{W} \vec{u}_i||\textbf{W}\vec{u}_k]))}
\end{equation}
where  means transposition and || is the concatenation operation.

\begin{equation}\label{key}
	\vec{u'}_i=\mathop{||}\limits^{H}_{h=1}\mathrm{ELU}(\sum _{j\in U_i} a^h_{ij}\textbf{W}^h\vec{u}_j)
\end{equation}
Where || represents concatenation, $a^{h}_{ij}$ are normalized attention coefficients calculated by the h-th attention mechanism and $\textbf{W}^h$ is the corresponding input liner transformation’s weight matrix. ELU is the activation function.

Finally, in the output layer of propagation structure attention networks, we replace the previous $concatenation$ with the average and use ReLU instead of ELU as the activation function. The process can be expressed as:
\begin{equation}\label{key}
	\vec{p}_i=\mathrm{ReLU}(\frac{1}{H}\sum^H_{h=1}\sum_{j\in U_i}a^h_{ij}\textbf{W}^h\vec{u'}_j)
\end{equation}

Through propagation structure attention networks, we get a representation vector of the propagation structure of news and the attention weight of each user node and its neighbour nodes.
\subsection{Prediction module}
The prediction module is a multi-layer feedforward neural network module. Based on the output of the text semantic attention networks and propagation structure attention networks, we use a softmax function in the output layer to predict the label of the Twitter news:
\begin{equation}\label{key}
	\hat{y}=\mathrm{softmax(ReLU}([V_t ||V_p]\textbf{W}_{tp}+b_{tp}))
\end{equation}
where || represents concatenation, $\textbf{W}_{tp}$ and $b_{tp}$ are parameters in the output layer and $V_t$ and $V_p$ are the output vectors of text semantic attention networks and propagation structure attention networks, respectively.

For each training process, we use the cross-entropy loss function to minimize the deviation between the predicted label and the real label:
\begin{equation}\label{key}
	\pounds(\Theta)=-ylog(\hat{y}_t -(1-y)log(1-\hat{y}_f))
\end{equation}
where $\Theta$ is the model parameter to be estimated, $y$ is the real label and $\hat{y}_t$ and $\hat{y}_f$ are the two labels predicted by the model: true and fake. 
\begin{figure*}[t]
	\centering
	\includegraphics[width=0.95\linewidth]{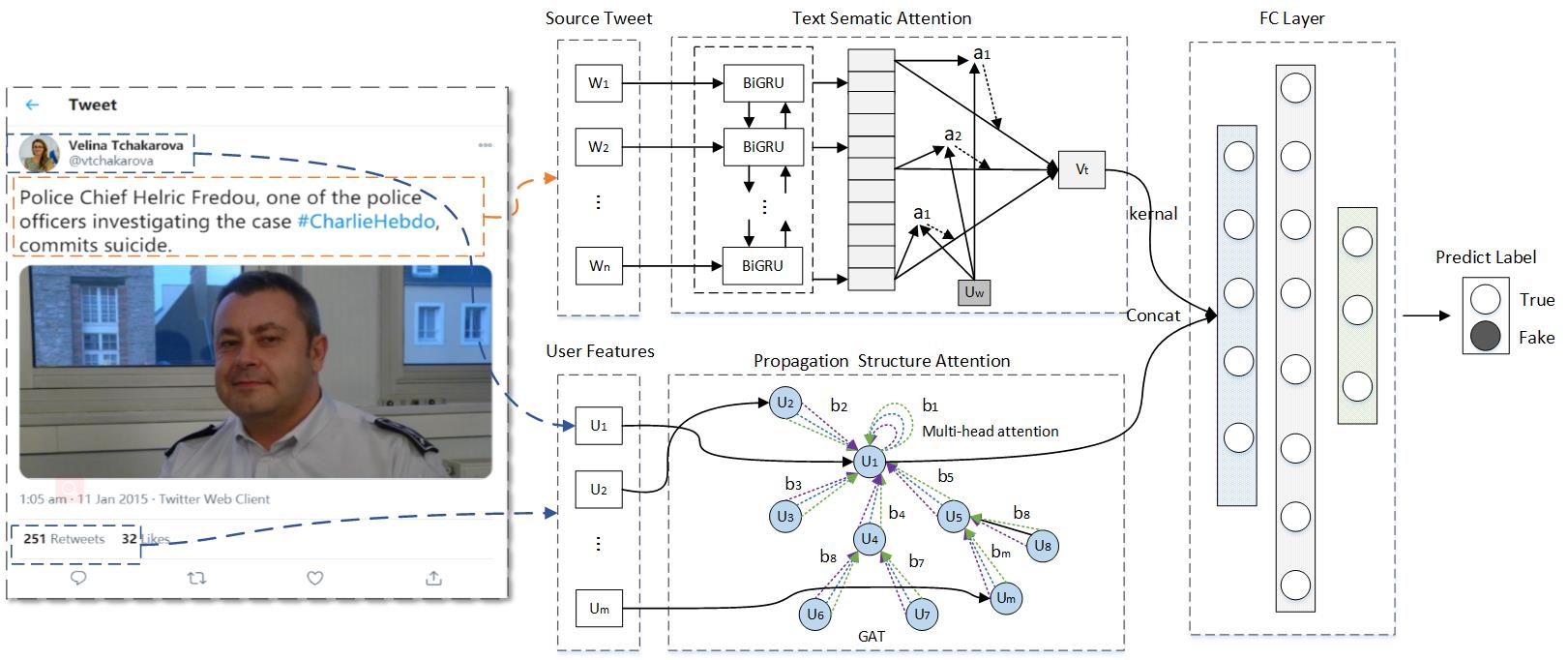}
	\caption{Architecture of the MVAN model. The two attention mechanisms represent two different kinds of information.}
	\label{fig:f2}
\end{figure*}
\section{Experiment and Results}

\begin{table}[]
	\caption{Statistics of the datasets.}
	\label{tab:my-table}
	\renewcommand\arraystretch{1.2}   
	\setlength{\tabcolsep}{2mm}{
		\begin{tabular}{|l|l|l|}
			\hline
			\textbf{Statistic} & \textbf{Twitter 15} & \textbf{Twitter 16} \\ \hline
			\# of users & 190,868 & 115,036 \\ \hline
			\# of source tweets & 742 & 410 \\ \hline
			\# of true & 372 & 205 \\ \hline
			\# of fake & 370 & 205 \\ \hline
			Avg. retweets per story & 292.2 & 308.7 \\ \hline
			Avg. words per source tweet & 13.25 & 12.81 \\ \hline
	\end{tabular}}
\end{table}

\begin{table}[h]
	\caption{A summary of user features in the two datasets.}
	\label{tab:my-table}
	\renewcommand\arraystretch{1.2}   
	\setlength{\tabcolsep}{3mm}{
		\begin{tabular}{|l|l|l|}
			\hline
			\textbf{No.} & \textbf{Features} & \textbf{Type} \\ \hline
			1 & user\_contributors\_enabled & Binary \\ \hline
			2 & user\_default\_profile & Binary \\ \hline
			3 & user\_default\_profile\_image & Binary \\ \hline
			4 & user\_favourites\_count & Integer \\ \hline
			5 & user\_follow\_request\_sent & Binary \\ \hline
			6 & user\_followers\_count & Integer \\ \hline
			7 & user\_following & Binary \\ \hline
			8 & user\_friends\_count & Integer \\ \hline
			9 & user\_geo\_enabled & Binary \\ \hline
			10 & user\_has\_extended\_profile & Binary \\ \hline
			11 & user\_listed\_count & Integer \\ \hline
			12 & user\_profile\_use\_background\_image & Binary \\ \hline
			13 & user\_protected & Binary \\ \hline
			14 & user\_statuses\_count & Integer \\ \hline
			15 & user\_verified & Binary \\ \hline
	\end{tabular}}
\end{table}

\subsection{Datasets}
Two well-known fake news datasets, Twitter15 and Twitter16\footnote{https://www.dropbox.com/s/7ewzdrbelpmrnxu/rumdetect2017.zip?dl=0}, were used to evaluate our MVAN model. The statistics of the database are shown in Table 2. Each dataset contains a collection of source tweets\footnote{Hashtags is a part of the textual information in the source tweet.}, along with their corresponding sequences of retweet user IDs. We chose only “true” and “fake” labels as the ground truth. Because the original dataset did not include user profiles, we used user IDs to crawl user feature information using Twitter API\footnote{https://dev.twitter.com/rest/public}. Some users have been deleted or abolished during the crawling process. In the Twitter15 and Twitter16 datasets, the total number of retweets is 190,868 and 115,036 respectively. The number of user information we collected through API is 177,049 and 108,801 respectively, and the corresponding percentages are 92.76\% and 94.58\%. It can be found that the missing users account for only a small part of the total. For missing user information, we use the mean value of other user features in the same propagation structure to fill it. Through the API, we crawled a total of 38 user features. Based on these previous works \cite{b1},\cite{b8},\cite{b47}, 15 common user features that are available on Twitter, which is summarized in Table 3.
\begin{table*}[]
	\caption{Comparison of multi-view attention networks with other models. The best model and the best competitor are highlighted in bold and underline. The results reported are the average of 10 runs. Note: The results of the model GCAN in the table are directly taken from the results shown in the original paper.}
	\scalebox{1}{
		\renewcommand\arraystretch{1.2}
		\setlength{\tabcolsep}{4.3mm}{
			\label{tab:my-table}
			\begin{tabular}{|l|l|l|l|l|l|l|l|l|}
				\hline
				\multirow{2}{*}{\textbf{Method}}& \multicolumn{4}{c|} {\textbf{Twitter 15}} & \multicolumn{4}{c|}{\textbf{Twitter 16}} \\
				\cline{2-9}
				&Acc  &Pre  &Rec  &F1  &Acc  &Pre  &Rec &F1  \\ \hline
				SVM-BOW & 0.6694 & 0.6357 & 0.6243 & 0.6292 & 0.6786 & 0.6783 & 0.6725 & 0.6747 \\ \hline
				BiLSTM & 0.7870 & 0.7370 & 0.780 & 0.7650 & 0.7850 & 0.7510 & 0.7730 & 0.7620 \\ \hline
				TextCNN & 0.7950 & 0.7630 & 0.7910 & 0.7770 & 0.7940 & 0.7370 & 0.7770 & 0.7590 \\ \hline
				CSI & 0.8573 & 0.8435 & 0.8596 & 0.8517 & 0.8465 & 0.8360 & 0.8557 & 0.8458 \\ \hline
				CRNN & 0.8626 & 0.8564 & 0.8644 & 0.8604 & 0.8708 & 0.8674 & 0.8669 & 0.8672 \\ \hline
				dEFEND & 0.8721 & 0.8669 & 0.8745 & 0.8707 & 0.8802 & 0.8785 & 0.8789 & 0.8787 \\ \hline
				GCAN & 0.8767 & 0.8257 & 0.8295 & 0.8250 & 0.9084 & 0.7594 & 0.7632 & 0.7593 \\ \hline
				G-SEGA& 0.8928 & 0.8914 & 0.8990 & 0.8952 & 0.9162 & 0.9034 & 0.9167 & 0.9093 \\ \hline
				MVAN& \textbf{0.9234} & \textbf{0.9264} & \textbf{0.9159} & \textbf{0.9224} & \textbf{0.9365} & \textbf{0.9324} & \textbf{0.9348} & \textbf{0.9336} \\ \hline
	\end{tabular}}}
\end{table*}

\subsection{Experimental Setup}
\textbf{Model comparison}: We compare our proposed models with other models, including some of the current state-of-the-art models.
\begin{itemize}
	\item \textbf{SVM-BOW} \cite{b42}: an SVM classifier using bag-of-words and N-gram (e.g., 1-gram, bigram and trigram) features (Ma et al., 2018). 
	
	\item \textbf{BiLSTM} \cite{b43}: a bidirectional RNN-based tweet model that considers the bidirectional contexts between targets and tweets (Augenstein et al., 2016).
	\item \textbf{TextCNN} \cite{b44}: a convolutional neural network model for obtaining the representation of each tweet and their classifications using a softmax layer (Chen et al., 2017). 
	\item \textbf{CSI} \cite{b45}: a state-of-the-art fake news detection model that captures the temporal pattern of user activities and scores users based on their behaviour (Ruchansky et al., 2017). 
	\item \textbf{CRNN} \cite{b46}: A Hybrid Deep Model for Fake News Detection]: A model combining RNN and CNN, which learns the text representations through the characteristics of users in the Twitter propagation path (Liu and Wu, 2018).
	\item \textbf{dEFEND} \cite{b22}:  a state-of-the-art co-attention-based fake news detection model that learns the correlation between the source article’s sentences and user profiles (Shu et al., 2019).
	
	\item \textbf{GCAN} \cite{b28}:  a state-of-the-art model that uses graph-aware co-attention networks (GCAN) to detect fake news (Lu et al., 2020). 
	
	\item \textbf{G-SAGE} \cite{b26}: a state-of-the-art detecting fake news by modeling conversation structure as a graph using GraphSAGE and BiLSTM (Li et al., 2020).
	
	\item \textbf{MVAN}: Our new deep neural network model, which uses both text semantic attention and propagation structure graph attention to detect fake news.
\end{itemize}

\textbf{Parameter setting}: In the text processing stage, we first cleaned the text information by removing useless expressions and symbols, uniform case, etc. We used GoogleNews pre-trained word2vec data with 300 dimensions for word embedding and set the maximum vocabulary to 250,000. The hidden size of BiGRU is 300, and the number of layers is 2. The batch size of the model was 64. The 15 user features shown in Table 3 were used as input data in the training process of the propagation structure attention network. We used 5-head attention in the graph attention network, and the number of graph attention network layers is 2. Moreover, we used the LeakyReLU nonlinearity with a negative input slope a = 0.3. In the training phase, we used Adam with a 0.001 learning rate to optimize the model, with the dropout rate set to 0.5. 

\textbf{Evaluation metrics}: For a fair comparison, we adopted the same evaluation metrics used in previous work. Therefore, the Accuracy, Precision Recall, and F1-measure (F1) were adopted for evaluation, which as described in the following equations:
\begin{equation}\label{key}
	Accuracy=\dfrac{TP+TN}{TP+TN+FP+FN}
\end{equation}
\begin{equation}\label{key}
	Precison=\dfrac{TP}{TP+FP}
\end{equation}
\begin{equation}\label{key}
	Recall=\dfrac{TP}{TP+FN}
\end{equation}
\begin{equation}\label{key}
	F1\_Measure=\dfrac{2\ast Recall\ast Precision}{Recall+Precision}
\end{equation}
where $TP$ are the true positive, $TN$ are the true negative, $FP$ the false positive and $FN$ the false negative predictions.

We followed GCAN \cite{b28} to split the datasets. In this paper, 70\% of the data were randomly selected for training and the remaining 30\% is used for testing. The results of the experiment are an average of ten times.
\begin{table*}[]
	\caption{Test accuracy on rumor detection with different models. Each model we ran 10 times and report the mean ± standard deviation. Our model significantly outperforms all the baseline based on t-tests(Confidence Level: 0.90, 0.95, 0,98). }
	\scalebox{1}{
		\renewcommand\arraystretch{1.2}
		\setlength{\tabcolsep}{2mm}{
			\begin{tabular}{|l|l|l|l|l|l|l|}
				\hline
				\multirow{2}{*}{\textbf{Method}}& \multicolumn{3}{c|} {\textbf{Twitter 15}} & \multicolumn{3}{c|}{\textbf{Twitter 16}} \\ 
				\cline{2-7} 
				& 90\% & 95\% & 98\% & 90\% & 95\% & 98\% \\ \hline
				SVM-BOW & 0.6694 ± 0.052 & 0.6694 ± 0.062 & 0.6694 ± 0.073 & 0.6786 ± 0.069 & 0.6786 ± 0.083 & 0.6786 ± 0.098 \\ \hline
				BiLSTM & 0.7870 ± 0.045 & 0.7870 ± 0.054 & 0.7870 ± 0.064 & 0.7850 ± 0.061 & 0.7850 ± 0.073 & 0.7850 ± 0.086 \\ \hline
				TextCNN & 0.7950 ± 0.044 & 0.7950 ± 0.053 & 0.7950 ± 0.063 & 0.7940 ± 0.060 & 0.7940 ± 0.071 & 0.7940 ± 0.085 \\ \hline
				CSI & 0.8573 ± 0.038 & 0.8573 ± 0.046 & 0.8573 ± 0.055 & 0.8465 ± 0.053 & 0.8465 ± 0.064 & 0.8465 ± 0.076 \\ \hline
				CRNN & 0.8626 ± 0.038 & 0.8626 ± 0.045 & 0.8626 ± 0.054 & 0.8708 ± 0.050 & 0.8708 ± 0.059 & 0.8708 ± 0.070 \\ \hline
				dEFEND & 0.8721 ± 0.037 & 0.8721 ± 0.044 & 0.8721 ± 0.052 & 0.8802 ± 0.048 & 0.8802 ± 0.057 & 0.8802 ± 0.068 \\ \hline
				GCAN & 0.8767 ± 0.036 & 0.8767 ± 0.043 & 0.8767 ± 0.051 & 0.9084 ± 0.043 & 0.9084 ± 0.051 & 0.9084 ± 0.061 \\ \hline
				G-SEGA & 0.8928 ± 0.034 & 0.8928 ± 0.041 & 0.8928 ± 0.048 & 0.9162 ± 0.041 & 0.9162 ± 0.049 & 0.9162 ± 0.058 \\ \hline
				MVAN & 0.9234 ± 0.029 & 0.9234 ± 0.035 & 0.9234 ± 0.041 & 0.9365 ± 0.036 & 0.9365 ± 0.043 & 0.9365 ± 0.051 \\ \hline
	\end{tabular}}}
\end{table*}
\begin{figure}[]
	\centering
	\includegraphics[width=0.99\linewidth]{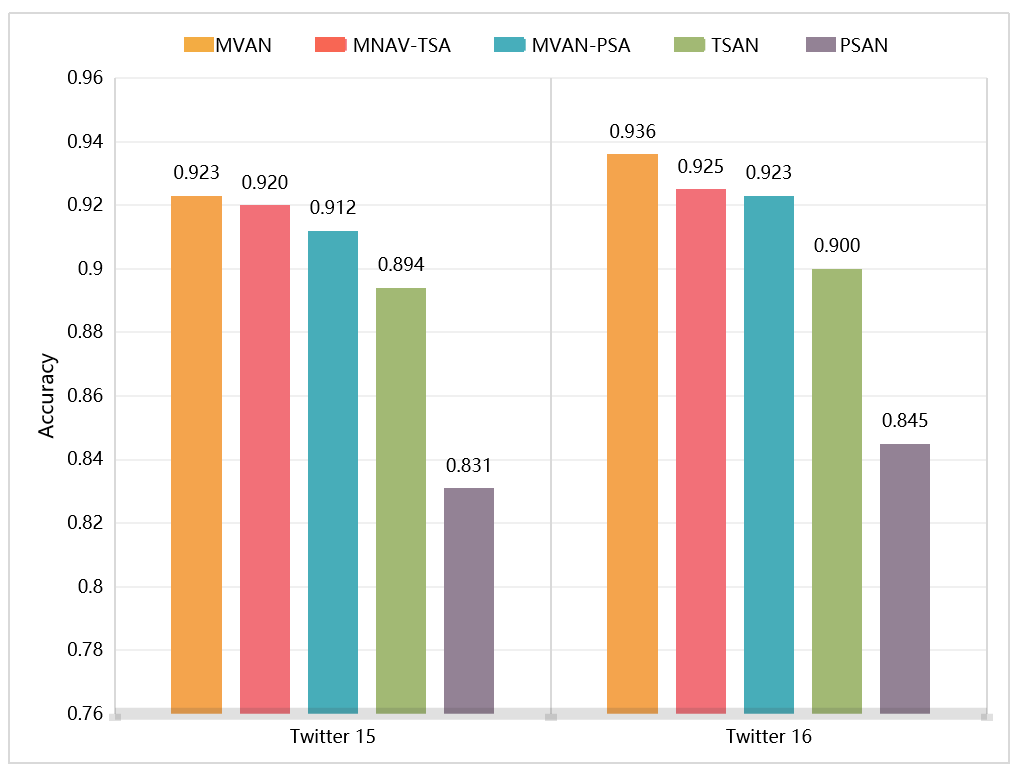}
	\caption{MVAN ablation analysis in accuracy. The results reported are the average of 10 runs.}
	\label{fig:3}
\end{figure}
\subsection{Experimental Results}
\begin{figure}[]
	\centering
	\includegraphics[width=0.99\linewidth]{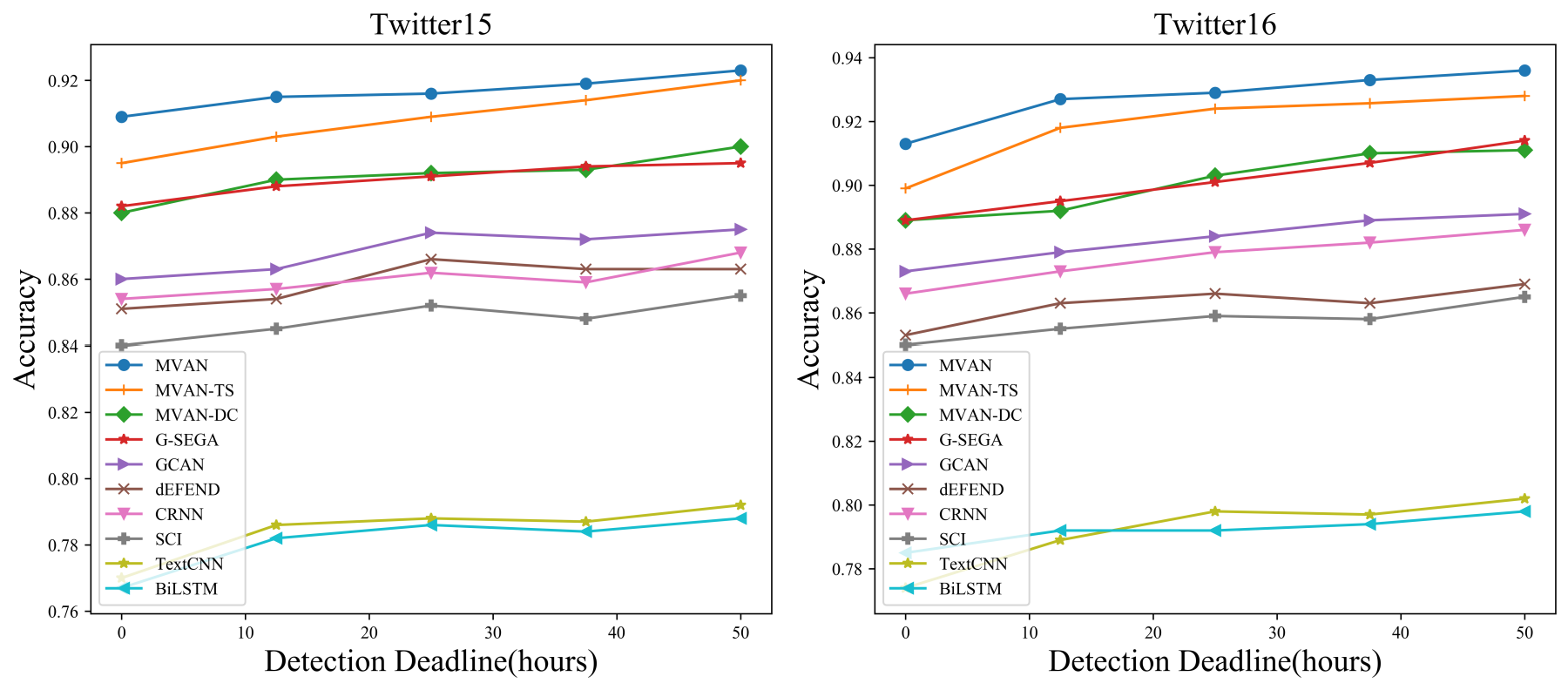}
	\caption{Early fake news detection results.}
	\label{fig:4}
\end{figure}
\begin{figure*}[t]
	\centering
	\includegraphics[width=1\linewidth]{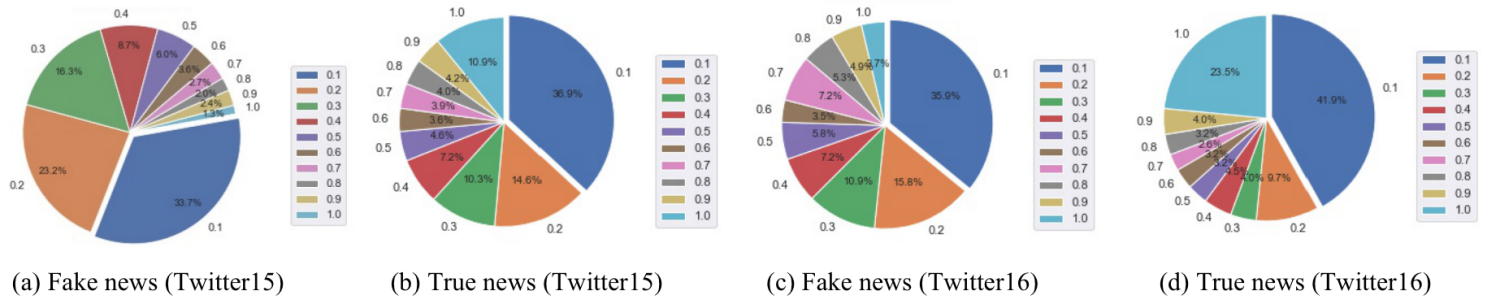}
	\caption{Distribution of statistics of word attention weight value.}
	\label{fig:5}
\end{figure*}
The main experimental results are shown in Table 4. The proposed MAVN model is significantly better than the state-of-the-art model in all the evaluation criteria in the two public datasets. Compared with the state-of-the-art model G-SEGA, the accuracy of Twitter15 and Twitter16 is improved by about 3.06\% and 2.03\%, respectively. Compared with the machine learning model SVM-BOW, the accuracy of our proposed model on both Twitter15 and Twitter16 datasets is improved by about 25\%. As shown in Table 4, we performed statistical tests on all models, and reported the average accuracy ± standard deviation with confidence levels of 0.90, 0.95, and 0.98 respectively. The results in Table 5 show that our model has a significant performance improvement on the two public datasets. MVAN can better represent text and propagation structural information, thereby improving the accuracy of fake news detection.

\subsection{Ablation Study}
To determine the relative importance of each module of the MVAN, we conducted a series of ablation studies on key parts of the model. The brief introduction of each model used for ablation research is as follows: 

\begin{itemize}
	\item \textbf{MVAN-TSA}: Both text and propagation structure information are used, but the text semantic attention mechanism is removed from the original MVAN model, and only BiGRU is used to encode the text.
	
	\item \textbf{MVAN-PSA}: Both text and propagation structure information are used but the MVAN model removes propagation structure attention mechanism and Node2vec is used to encode the propagation graph structure directly. 
	
	\item \textbf{TSAN}: Text semantic attention network only uses text data to classify news.
	
	\item \textbf{PSAN}: The propagation structure attention network only uses propagation structure data to classify news.
	
\end{itemize}
The experimental comparison results are shown in Fig. 3. We found that when the MVAN model removed the text semantic attention mechanism or propagation structure attention mechanism, the performance dropped by about 1\%. This shows that these two attention mechanisms have a certain effect on our model performance. When the model used TSAN only, the performance of the model dropped by 2.9\% to 3.6\%, because the model loses very important propagation structure information. In addition, if only PSAN was used, the performance of the model dropped by about 9\% on both data sets, because the model does not even read the text content of the news itself. However, the performance of PSAN on Twitter15 and Twitter16 data sets reached 8.31\% and 8.45\%, respectively, which proves that numerous clues can be used to detect fake news in the propagation structure.
\subsection{Early detection performance}
It is very important to detect fake news in the early stages of propagation so that preventive measures can be taken as quickly as possible. In the early detection tasks, all user information after the detection deadline is invisible during the test. The earlier the deadline, the less propagation information is available. 

While comparing the previous comparison models, we add ST-GCN \cite{b48} and DCRNN \cite{b49} models for comparison. These two models are neural network models for processing temporal sequence graphs. For the convenience of comparison, we replaced the propagation structure attention modules in our model with ST-GCN and CDRNN, which we named MVAN+ST and MVAN+DC, respectively. As can be seen in Figure 4, our model can achieve an accuracy of approximately 91\% in the earliest stage. DCRNN is a combination of GCN and RNN. It only supports the timing input of the isomorphic graph model. Therefore, we build a corresponding DCRNN model for each time window for testing, but because each graph corresponds to For each structure, the RNN module in the model training process is equivalent to only one output, so the performance of this model is not fully utilized. ST-GCN can be used to convolve information in both spatial and temporal dimensions at the same time. Therefore, the effect of ST-GCN is quite good, and there is a great possibility of improvement afterwards. Moreover, from the broken line diagram, the curve of MVAN is very stable, indicating that our model has good robustness and high performance in early fake news detection.
\subsection{Interpretability analysis}
\begin{figure}[t]
	\centering
	\includegraphics[width=1\linewidth]{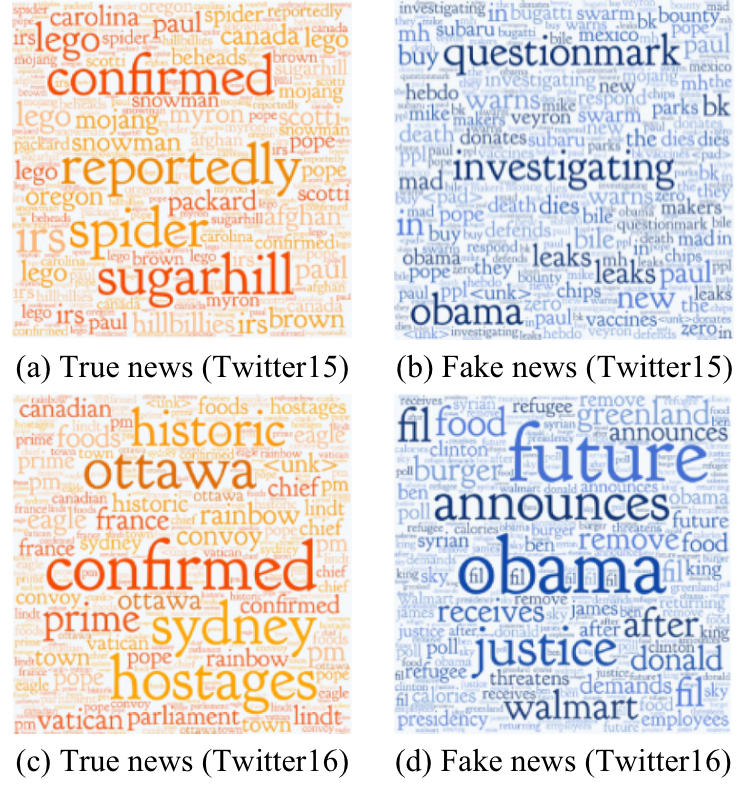}
	\caption{Evidence words through word cloud visualization. A larger font indicates a higher text semantic attention weight.}
	\label{fig:6}
\end{figure}
\begin{figure*}[]
	\centering
	\includegraphics[width=0.98\linewidth]{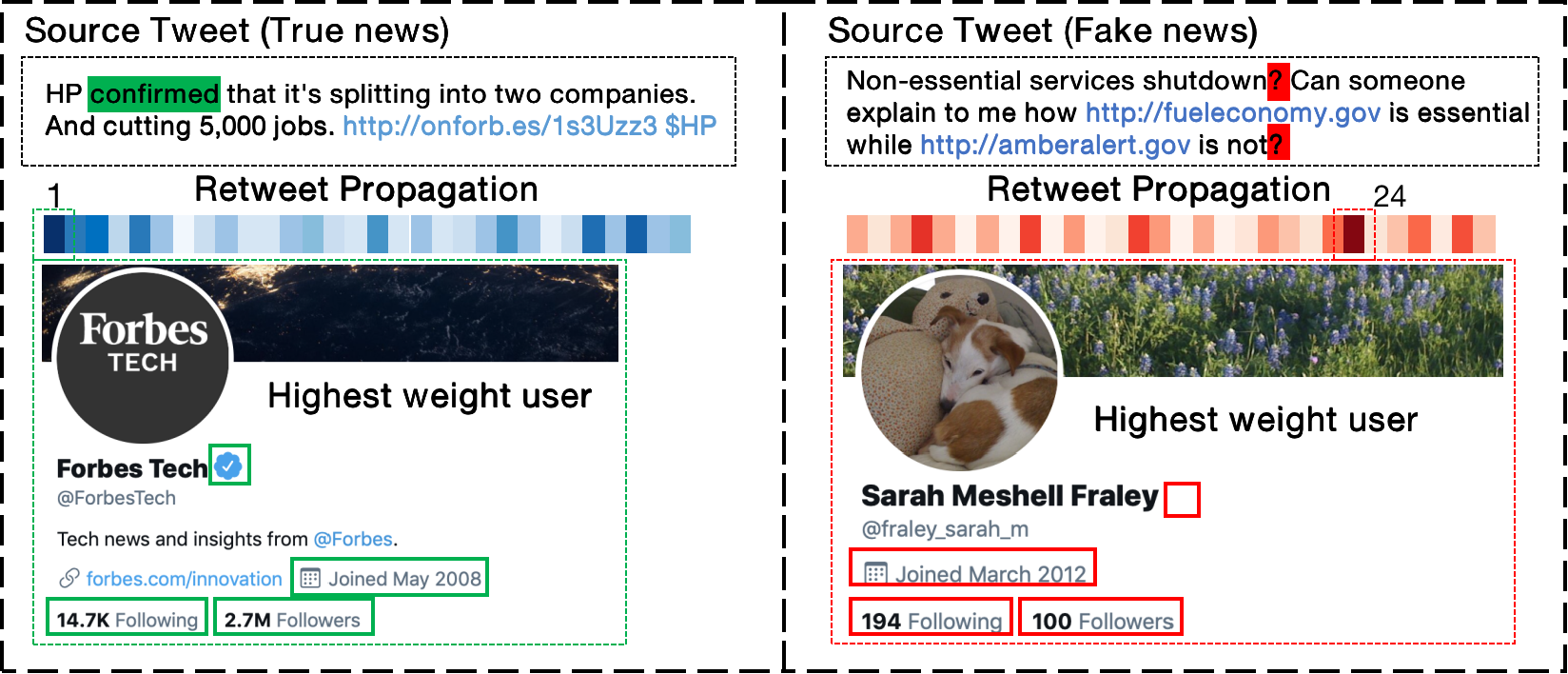}
	\caption{Real case analysis of true news and fake news. The key clue words in the source tweet are highlighted by text semantic attention weights. Visualization of attention weights for retweet propagations of a fake and a true source tweets. From left to right is retweet order, and dark colors refer to higher attention weights.  The user with the highest weight in the retweet propagation is marked, and the main characteristics of the user are displayed.
	}
	\label{fig:7}
\end{figure*}
The text semantic attention assigns relative weight values to the words in the source tweet (the weight value is between 0 and 1). We analyzed the weight distribution of the words in the two datasets Twitter15 and Twitter16. Regardless of whether it is fake news or true news, most words had a small weight of about 0.1 (Fig. 5), implying that most words have little effect in distinguishing the truth of the news. Additionally, we found that the proportion of words with an attention weight equal to 1 in real news was much higher than that in fake news. For example, in Twitter15 and Twitter16, words with an attention weight of 1 accounted for 1.3\% and 3.7\%, respectively, in fake news and 10.9\% and 23.5\%, respectively, in true news.

Therefore, according to the word weights given by the text semantic attention mechanism in the model, we visualized high-weight words in the two data sets with word clouds. These words had more weight on real news, “confirmed”, “Sugarhill”, “reportedly”, “hostages”, etc. (Fig. 6). However, in the fake news, “question mark”, “investigating”, “Obama”, “announces”, “future ”, etc., these words were more weighted. Furthermore, true news contains more certain and authoritative words, while fake news may be vague and suspicious.

To further analyze the interpretability of the model, we selected two examples from the real data set for analysis. As shown in Fig. 6, the left one is true news and the right one is fake news. We used the text semantic attention mechanism to highlight the evidence words. We thought that if the word “confirmed” was included in the news, then the news is more likely to be true. Contrarily, if there were more “?” in the news, which represents doubtful information, then the news may be fake. 

In addition, we used the propagation structure attention mechanism to find the most weighted key clue retweet users in the propagation structure. We randomly select a fake and a real source tweet, and draw their weights according to the propagation structure attention, as shown in Figure 7, where the horizontal direction from left to right represents the order of retweets. As shown in the figure 7, in the retweet propagation structure of the real news, the first user has the highest weight, and the 24th user of the fake news has the highest weight. The results show that to determine whether a news is fake, one should first check the features of users who early retweet the source tweet. In the propagation structure of fake news, user weights are more evenly distributed. The features of key Twitter users in the propagation structure of true news and fake news were markedly different (Fig. 7). The user accounts in the true news were created earlier, with authentication icons, profiles and followed by many users and were followed by many users. Such users are generally more authoritative official news accounts. However, the user accounts in fake news are created late, with no certification, no profile and without many followers. General, such a user account perhaps spreads fake news.
\section{Conclusion}
We propose a new deep learning model for fake news detection, MVAN, which combines two attention mechanisms, text semantic attention and propagation structure attention, to simultaneously capture the important hidden clues and information in the source tweet text and the propagation structure. The evaluation results using two public data sets show that MVAN has strong performance and reasonable interpretation ability. In addition, MVAN can provide early detection of fake news with satisfactory performance. We believe that MVAN can be used not only for fake news detection but also for other text classification tasks on social media, such as sentiment classification, topic classification, insult detection, etc.

In future work, the users’ reply information will be added to further improve the performance of the model. Subsequently, GNN combined with the attention mechanism will be used to capture the key information hidden in the conversation structure graph composed of the source tweet and its replies. We will start from a real-world perspective and detect fake news from more different perspectives. 
\subsubsection{Acknowledgments.}
This  work  was  partially  supported  by the project MOST 109-2221-E-006-173 funded by Ministry of Science and Technology, Taiwan.

\end{document}